\documentclass[conference]{IEEEtran}
\IEEEoverridecommandlockouts
\usepackage{cite}
\usepackage{amsmath,amssymb,amsfonts}
\usepackage{algorithmic}
\usepackage{graphicx}
\usepackage{textcomp}
\usepackage{graphics} 
\usepackage{epsfig} 
\usepackage{mathptmx} 
\usepackage{times} 
\usepackage{amsmath} 
\usepackage{amssymb}  
 \usepackage[table]{xcolor}
\usepackage{hyperref}
\usepackage{cleveref}
\usepackage{booktabs}
\usepackage{tabularx}
\usepackage{float}
\def\BibTeX{{\rm B\kern-.05em{\sc i\kern-.025em b}\kern-.08em
    T\kern-.1667em\lower.7ex\hbox{E}\kern-.125emX}}

\IEEEoverridecommandlockouts

\begin{document}

\title{Supervised Machine Learning based Ensemble Model for Accurate Prediction of Type 2 Diabetes}

\author{\IEEEauthorblockN{1\textsuperscript{st} Ramya Akula}
\IEEEauthorblockA{\textit{Industrial Engineering} \\
\textit{University of Central Florida}\\
Orlando, USA \\
ramya.akula@knights.ucf.edu}
\and
\IEEEauthorblockN{1\textsuperscript{st} Ni Nguyen}
\IEEEauthorblockA{\textit{Industrial Engineering} \\
\textit{University of Central Florida}\\
Orlando, USA \\
nithi@knights.ucf.edu}
\and
\IEEEauthorblockN{2\textsuperscript{nd} Ivan Garibay}
\IEEEauthorblockA{\textit{Industrial Engineering} \\
\textit{University of Central Florida}\\
Orlando, USA \\
igaribay@ucf.edu}
}

\maketitle

\begin{abstract}
According to the American Diabetes Association(ADA), 30.3 million people in the United States have diabetes, but only 7.2 million may be undiagnosed and unaware of their condition. Type 2 diabetes is usually diagnosed for most patients later on in life whereas the less common Type 1 diabetes is diagnosed early on in life. People can live healthy and happy lives while living with diabetes, but early detection produces a better overall outcome on most patient's health. Thus, to test the accurate prediction of Type 2 diabetes, we use the patients' information from an electronic health records company called Practice Fusion, which has about 10,000 patient records from 2009 to 2012. This data contains individual key biometrics, including age, diastolic and systolic blood pressure, gender, height, and weight. We use this data on popular machine learning algorithms and for each algorithm, we evaluate the performance of every model based on their classification accuracy, precision, sensitivity, specificity/recall, negative predictive value, and F1 score. In our study, we find that all algorithms other than Naive Bayes suffered from very low precision. Hence, we take a step further and incorporate all the algorithms into a weighted average or soft voting ensemble model where each algorithm will count towards a majority vote towards the decision outcome of whether a patient has diabetes or not. The accuracy of the Ensemble model on Practice Fusion is 85\%, by far our ensemble approach is new in this space. We firmly believe that the weighted average ensemble model not only performed well in overall metrics but also helped to recover wrong predictions and aid in accurate prediction of Type 2 diabetes. Our accurate novel model can be used as an alert for the patients to seek medical evaluation in time.
\end{abstract}


\section{Introduction}
In 2017, Center for Disease Control(CDC) \footnote{\href{https://www.cdc.gov/diabetes/pdfs/data/statistics/national-diabetes-statistics-report.pdf}{Center for Disease Control National Diabetes Statistics Report, 2017.}} estimates 84 million adult Americans have a prediabetic condition, and 90\% of them go undiagnosed. In 2018, according to the American Diabetes Association(ADA) \footnote{\href{http://www.diabetes.org/diabetes-basics/statistics/?referrer=https://www.google.com/}{American Diabetes Association Statistics, 2018.}}, 30.3 million people in the United States have diabetes, but 7.2 million may be undiagnosed and unaware of their condition. About 1.5 million new cases of diabetes are diagnosed in the United States every year. Type 2 diabetes is usually diagnosed for most patients later on in life whereas the less common Type 1 diabetes is diagnosed early on in life for most patients with the peak age usually being around 14. People can live healthy happy lives while living with diabetes, but early detection produces a better overall effect for most patients’ health. Initial onset of Type 2 diabetes can produce mild to no symptoms, so many patients may not be diagnosed until 7-10 years after onset. The uncontrolled high blood sugar damages nerves over time. If left untreated, diabetic patients face an increased risk of diabetic retinopathy which in turn leads to blindness, kidney disease leading to renal failure, and nerve disease leading to neuropathy and numbness in extremities among other serious conditions. In this work, to test the accurate prediction of Type 2 diabetes, we started with raw data from an electronic health records company called Practice Fusion and about 10,000 patient records from 2009 to 2012. We chose to focus on this specific biometrics because there has been significant evidence to show this may affect the determination of diabetes:
\begin{itemize}
    \item \textbf{Age:} According to 2017 CDC report, the rate of diagnosed diabetes increased with age. Among adults ages 18-44, 45-64 years and 65 years and older,  of about 4\%, 17\%, and 25\% had diabetes. Further, according to 2018 ADA report, an estimated 11.7\% of women (with a 95\% C.I.) have diabetes over the age of 18. A rated 12.7\% of men (with a 95\% C.I.) have diabetes over the age of 18. 
    \item \textbf{Blood Pressure:} According to 2017 CDC report, 73.6\% of those living with diabetes have a systolic blood pressure of 140 mm Hg or higher and diastolic blood pressure of 90 mm Hg or higher, and they were on prescription medication for high blood pressure. It is a high-risk factor of living with diabetes as it could lead to cardiovascular problems.
    \item \textbf{Height, Weight-BMI:} According to National Institute of Diabetes and Digestive and Kidney Diseases \footnote{\href{https://www.niddk.nih.gov/health-information/diabetes/overview/what-is-diabetes/prediabetes-insulin-resistance}{Insulin Resistance and prediabetes from National Institute of Diabetes and Digestive and Kidney Diseases}}, experts believe that obesity, especially visceral fat, is the primary cause of insulin resistance.
\end{itemize}
Consequently, to test the accurate prediction of Type 2 diabetes, we build a supervised machine learning ensemble model\footnote{\href{https://github.com/akula01/Supervised-Machine-Learning-Ensemble-model-for-Type-2-Diabetes-Prediction.git}{Code and Datasets on GitHub Repository: Supervised Machine Learning Ensemble Model for Type-2 Diabetes Prediction}} using seven standard classification algorithms: k-Nearest Neighbors, Support Vector Machines, Decision Trees, Random Forest, Gradient Boosting, MLP Neural Network, and Naive Bayes. For each algorithm, we tune hyperparameters to produce the best accuracy, and at the end, we evaluate each model's performance based on their classification accuracy, precision, sensitivity, specificity/recall, negative predictive value, and F1 score.

Organization of this paper as follows: as Section 2 consists of related work in this field. Section 3 explains, the proposed approach along with problem formulation, datasets, and data preparation, and the components and description ensemble model. Section 4 describes the results and analysis of the developed ensemble model. This section illustrates the empirical study to determine the performance of every machine learning algorithm used. Section 5 briefly concludes the paper and followed by the future scope of extension.

\section{Related Work}
 Systematic literature conducted in \cite{kavakiotis2017machine} showed that machine learning and data mining tools are applied in the field of diabetes research concerning prediction and diagnosis, diabetic complications, genetic background and environment, and health care management. In general, 85\% of those characterize supervised learning approaches and 15\% by unsupervised ones, and more specifically, with the association of different rules. Support vector machines (SVM) is the most successful and widely used algorithm. A novel update on race/ethnic differences in children and adults with Type 1 diabetes, children with Type 2 diabetes in Latino sub-populations reviewed \cite{spanakis2013race}. Studies conducted in \cite{baier2004genetic}, \cite{alghamdi2017predicting}, \cite{sisodia2018prediction}, \cite{kandhasamy2015performance}  have attempted to produce a model for diabetes detection with machine learning. An empirical study conducted in \cite{espeland2007reduction} observed that at one year, intensive lifestyle intervention resulted in clinically significant weight loss in people with Type 2 diabetes. This weight loss associated with improved diabetes control and cardiovascular disease risk factors and reduced medicine use in intensive lifestyle intervention versus diabetes support and education. Continued intervention and follow-up will determine whether these changes are maintained and will reduce cardiovascular disease risk. Diabetes Mellitus is a metabolic disease where random blood glucose levels lead to the risk of many diseases like heart attack, kidney disease, and renal failure. Diabetes Mellitus diagnosed in \cite{krati2014diagnosis} uses only K- Nearest neighbor algorithm. On the other hand, diabetes complications are predicted in \cite{dagliati2018machine} by embedding machine learning algorithms into data mining pipelines, which can combine them with traditional statistical strategies, to extract knowledge from data. In this research, the missing information from the collected data is dealt with utilizing random forest and applying the proper approach to handle class imbalance and then used Logistic Regression with stepwise feature selection to predict the onset of retinopathy, neuropathy, or nephropathy, at different time scenarios. That is, at 3, 5, and 7 years from the first visit at the Hospital Center for Diabetes, but not from the diagnosis. This method helped in the smooth translation of model assessment to the clinical practice. The study conducted in \cite{kandhasamy2015performance} compared the performance of supervised machine learning algorithms that were used to predict diabetes. This study employed the Pima Indians data set and k Nearest Neighbors, Decision Trees, Random Forest, and SVM. Of the four classifiers, Decision Tree was able to achieve the highest accuracy of 73.82\%. After removing inconsistent/noisy data, they were able to achieve 100\% accuracy using the Random Forest algorithm or k Nearest Neighbors algorithm with the k=1 neighbor. However, there is no record of further performance tests. In 2018,\cite{sisodia2018prediction} used three different machine learning classification models such as Naive Bayes, Support Vector Machine, and Decision Tree to predict diabetes. In their study, they found that the Naive Bayes classification the highest accuracy of 76.30\%. Unlike the previous works that focused either particular classifier-set or a Pima Indians dataset that is heavily biased towards limited female population, we use a new approach and dataset, yet use the Pima Indians dataset for baseline comparison. That is, the ensemble method of machine learning is our approach to predict Type 2 diabetes more accurately and we use Practice Fusion Dataset. We firmly believe that the weighted average ensemble model not only performed well in overall metrics but also helped in recover wrong predictions. While the accuracy of the previous works on Pima Indians dataset was less than 80\%, the accuracy of our Ensemble model reached 89\% for the same dataset. For Practice Fusion dataset, due to different approaches of the winners of the Kaggle competition, we were unable to accomplish one to one comparison on model comparison unlike the previous dataset. However, the accuracy of the Ensemble model on Practice Fusion is 85\%, by far our ensemble approach is new in this space. Description of each of the classifiers, ensemble model and their performance on Practice Fusion dataset are detailed in the later sections, since that is the primary focus of this work. 
 
\section{Proposed Learning Approach}
\begin{figure*}[h!]
    \centering
    \includegraphics[scale=0.75]{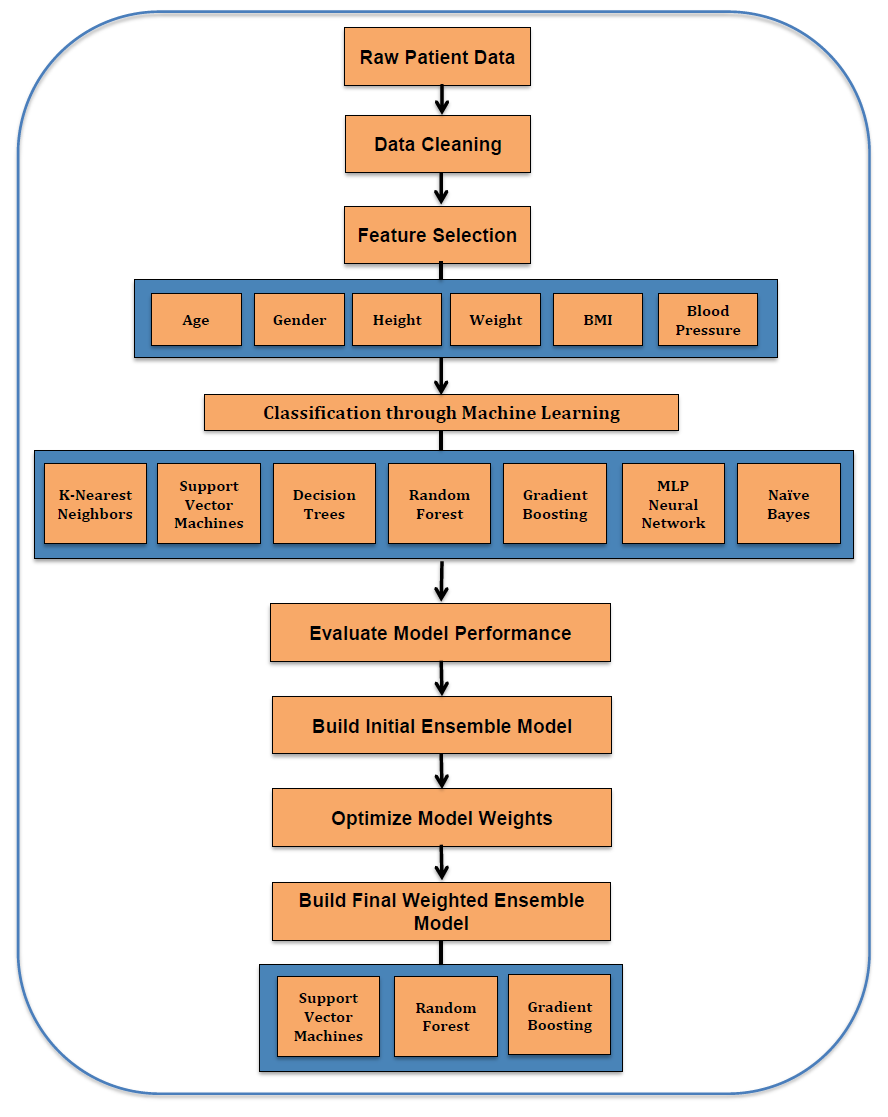}
    \caption{Block Diagram of Supervised Machine Learning Ensemble Model for Type 2 Diabetes Prediction}
    \label{fig:block}
\end{figure*}

We explored the data through seven machine learning algorithms such as k Nearest Neighbors, Support Vector Machines, Decision Trees, Random Forest, Gradient Boosting, Neural Network, and Naive Bayes. For each algorithm, we perform hyperparameter tuning and then assess the model’s performance with classification. Before performing ensemble learning, we will look at the classification report for each algorithm as shown in the block diagram in Fig \ref{fig:block}. 
\begin{figure*}[h!]
    \centering
    \includegraphics[scale=0.50]{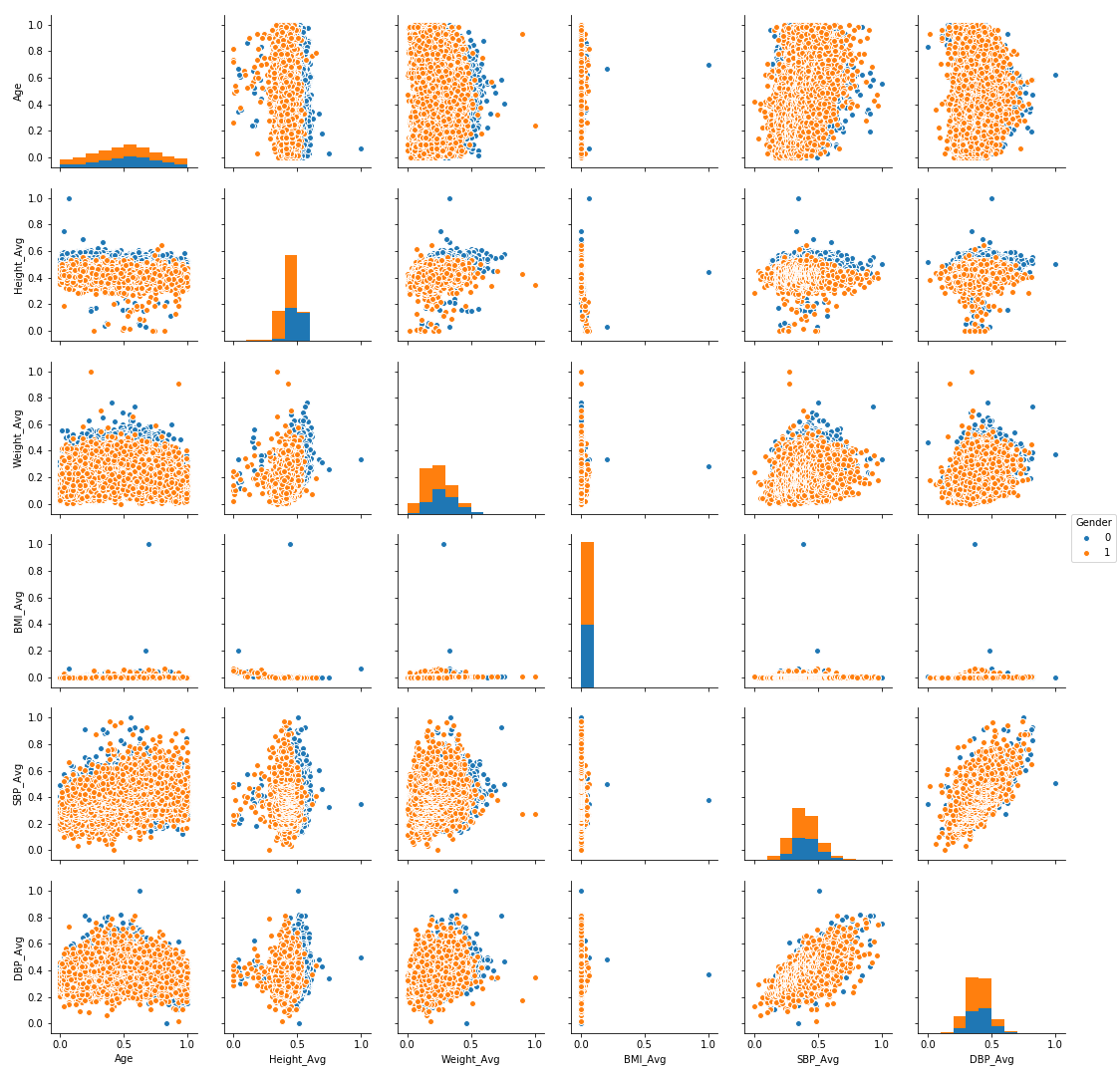}
    \caption{Pair plot of differing variables after the data normalization of Practice Fusion dataset. SBP = Systolic Blood Pressure, DBP = Diastolic Blood Pressure, Gender 1 = Female, Gender 0 = Male}
    \label{fig:pairplot}
\end{figure*}
In this section, we describe the data preparation for the standard machine-learning models.

\subsection{Data Preparation}
In 2012, Practice Fusion, an electronic medical health records company, released de-identified records for 9,948 patients across the United States on Kaggle in a competition to build a diabetes detection model. For those patients, records from years 2009 to 2012 include extensive information including patient vitals at each medical office visit; have taken medications, allergies, medical conditions, and lab results. From this, we decided to filter the data to focus on patient weight, height, age, gender, diastolic and systolic blood pressure, and BMI. For each patient in the data set, we replaced all zeros in the data with NaN to be able to calculate the mean with NaN mean function. From there, we estimate the minimum, maximum, and average or mean values of weight, height, BMI, and both diastolic and systolic blood pressure for each patient for all of their vitals took over those four years. So for each patient, we were comparing those 15 values in addition to their age and gender for a total of 17 values, as shown in Fig \ref{fig:pairplot}. There will also be one column that represents the outcome of whether the patient has diabetes or not. Our ultimate goal is to produce something that can help clinicians predict diabetes for their patients, and ultimately these vitals are the most convenient and prevalent to obtain for any patient. For a baseline comparison, we use Pima Indians that is used by previous works. This dataset is heavily biased towards female population and has about 800 records.
In particular for each algorithm we calculate the following performance measures along with the accuracy of each model, as accuracy is a hoax measure if used alone to test the performance of any machine learning model as shown in Table \ref{tab:class_perform_practice}.
\begin{itemize}
    \item \textbf{Sensitivity:} Sensitivity will refer to the ratio that for all patients that had diabetes, the percent the algorithm classified as diabetic.
    \item \textbf{Precision:} Precision will refer to the ratio that for all the predictions that a patient had diabetes, the percentage that was diabetic.
    \item \textbf{Negative Prediction:} Negative predictive value refers to the ratio of the predicted value of all patients being non-diabetic, the percentage that was non-diabetic.
    \item \textbf{Specificity:} Specificity refers to the ratio that for all patients that were non-diabetic, the percentage the algorithm classified as non-diabetic.
    \item \textbf{F1-Score:} F1 score is the harmonic mean of precision and recall. It maintains a good balance of positive predictive values and sensitivity.  
\end{itemize}

\section{Supervised Learning - Ensemble Model}
In this section, we explain the performance of every classifier used for this prediction task, along with a brief description of the utility of the mentioned standard classifiers.
\subsection{K-Nearest neighbor}
K-Nearest Neighbor(k-NN)\cite{keller1985fuzzy} is a type of classification algorithm that predicts the outcome of a data point based on comparison to its closest neighbors' outcomes. The k here stands for the number of neighbors that the model is comparing to. The k-Nearest neighbor algorithm will calculate the closest or nearest neighbors to a data point by comparing the data point's variable values to the values of other points. It does this by summing the distance of each variable in one data point and each variable in another point in the set. The distance, in this case, will be calculated with the Euclidean distance formula. The closest points will have the smallest sums in comparison to the data point. We compare our test point to k numbers of closest neighbors and depending on whether that neighbor is diabetic or not; it will affect the prediction of whether our point is also diabetic. 
\begin{figure*}[h!]
    \centering
    \includegraphics[scale=0.4]{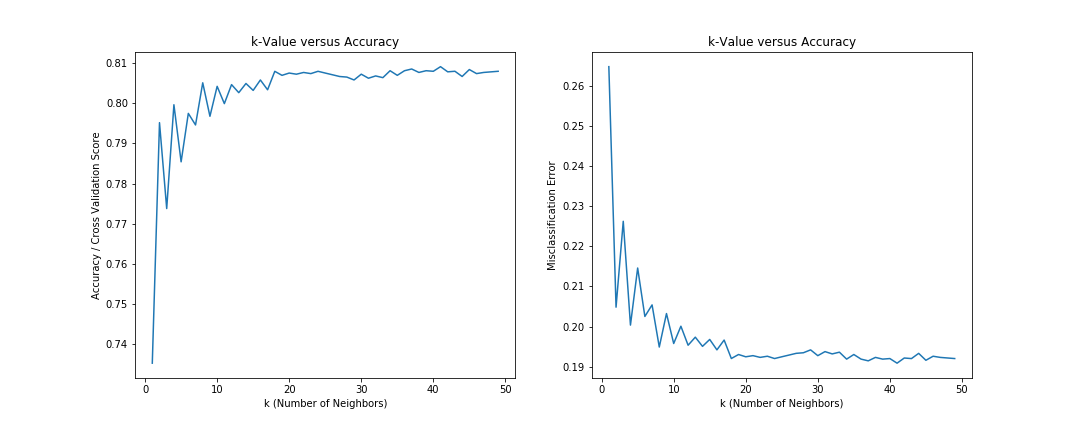}
    \caption{Cross Validation error for K nearest Neighbors}
    \label{fig:knncrossval}
\end{figure*}
For example, if the K number of neighbors is 3, we compare individual data point to the three closest data points to it. Those 3 points each have an equal vote in determining if the test point is diabetic or not. If for example, 2 of those points are diabetic and one is not diabetic, then our prediction lean towards the individual being diabetic based on a majority vote. Training and testing accuracy of KNN with the cross-validation of KNN is as shown in Fig \ref{fig:knncrossval}. After cross-validating the data with k Fold testing, we found that the optimal k neighbors to compare to was 41. The accuracy of the model was about 81.5\%,  but the precision of the model was very low at 6.1\%, negative predictive value was high at 98.8\%, sensitivity was 54\%, and specificity was 82.1\%. This result shows that when the algorithm predicted that a patient was non-diabetic, the algorithm was correct 98.8\% of the time. But, if a patient anticipated being diabetic, it was only right 6\% of the time. This result explains a low rate of false negatives but a high rate of false positives, which is better than the reverse as we instead a patient get tested for being diabetic than stating that they were non-diabetic and being wrong as that patient never get tested. F1 score for this model was 11\%.
 
\subsection{Support Vector Machines}
\begin{figure}[h!]
    \centering
    \includegraphics[scale=0.4]{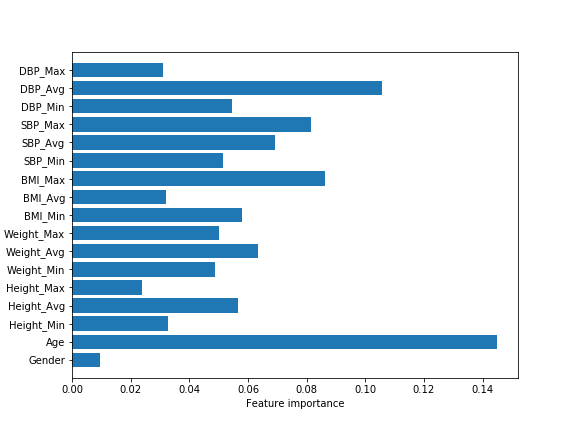}
    \caption{Feature Importance for Support Vector Machine.}
    \label{fig:svmfeat}
\end{figure}

Support Vector Machine(SVM)\cite{cortes1995support} is another machine learning algorithm used towards classification. A support vector machine will divide a dataset into two classes, in our case diabetic and non-diabetic, by separating it with the hyperplane that best divides them. We consider a 2D data set with only data points and one variable like gender being compared. SVM creates a line/plane of the division to split the dataset so that each gender is separated. Points closer to the plane will affect the position of the plane. Once the plane has been placed, we are more confident that points further from the plane were correctly classified while points closer to the boundary, that is less confident on correct classification. As we add more and more variables for comparison, this plane becomes a hyperplane of higher and higher dimensions to segregate our data better. The importance of features is shown in Fig \ref{fig:svmfeat}.
After normalizing our data, we performed parameter tuning and found that a linear SVM with C of 1. The accuracy of the model was about 81.1\%, but the precision of the model was extremely low at 0.5\%, negative predictive value was high at 99.6\%, sensitivity was 25\%, and specificity was 81.4\%. Again, like the previous algorithm, this shows that when our predicted value, i.e., a patient was non-diabetic, the algorithm was correct almost 100\% of the time. But, if a patient was predicted as diabetic, almost never correct. F1 score for this algorithm was 1.05\%.

\subsection{Decision Tree}
Decision Tree\cite{kohavi1996scaling} is a supervised machine learning algorithm also used towards classification. The classification model is built in a tree structure. 
\begin{figure}[h!]
    \centering
    \includegraphics[scale=0.4]{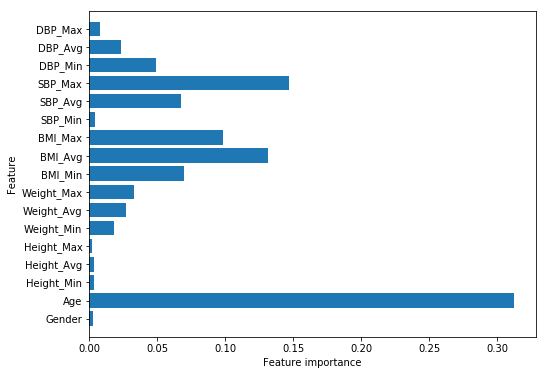}
    \caption{Feature Importance for Decision Tree}
    \label{fig:treefeat}
\end{figure}
The data set is broken down into smaller subsets that become nodes and inter nodes in the tree structure. Each split is formed on a particular variable. Each decision or root node can have two or more branches, and each leaf node will represent classification.
\begin{figure}[h!]
    \centering
    \includegraphics[scale=0.3]{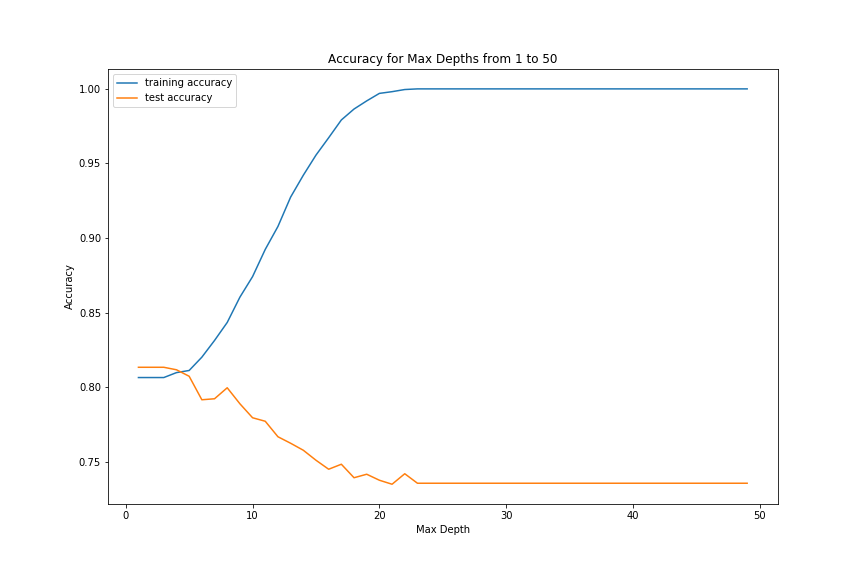}
    \caption{Accuracy of Decision Tree}
    \label{fig:decisionacc}
\end{figure}
After parameter tuning, we found that decision tree with a max tree depth of 7 produced the best accuracy for the model at  79.2\%. While the feature importance for this algorithm is shown in \ref{fig:treefeat} Accuracy is shown in Fig \ref{fig:decisionacc}. The precision of the model was still low but nowhere nearly as low as the previous models at 23.5\%, negative predictive value was high at 92.0\%, sensitivity was 40.3\%, and specificity was 84\%. Compared to previous models, we gained quite a bit of precision, sensitivity, and specificity but lost some of our negative predictive value. F1 score for this algorithm was 29.7\%.
 
\subsection{Random Forest}
Random Forest\cite{liaw2002classification} is an ensemble machine learning algorithm used for classification. It starts with and uses decision trees like previously, however, the model creates an entire forest of random uncorrelated trees to arrive at the classification decision. Multiple trees are created from multiple training sets generated randomly with replacement. For each tree, node splits are stopped by choosing only a random subset of features at each split. This prunes each tree and produces trees that are random and uncorrelated. Then when the model makes a prediction, it will take the average of all decisions that the individual trees in the forest make to come to a final classification decision. After parameter tuning, we found that a max tree depth of 9 and n-estimators of 250 trees in the forest for decision making produced the most optimal result. 

The accuracy of the model was about 82.4\%. The precision of the model was 12.4\%, negative predictive value was high at 98.4\%, sensitivity was 64\%, and specificity was 83.0\%. This model produced the highest specificity so far, which combined with the negative predictive value shows that when this model predicts that a patient is non-diabetic, that is true 98.4\% and of the non-diabetic patients, 83\% were classified correctly. F1 score for this algorithm was 20.7\%.
 
\subsection{Gradient Boosting}
Gradient Boosting{friedman2001greedy} is another ensemble machine learning algorithm used for classification that uses boosting to improve performance. We start by fitting an initial decision tree to the data. After the first tree is evaluated, the next tree is constructed to cover the difference between the first tree and the actual target. It does so by reconstructing the residual. This means that when combined with the previous tree, it will minimize the overall prediction error. The next tree will continue to do so with all previous trees so that in the end each subsequent tree is optimizing the differentiable loss function and taking a further step in the direction that minimizes the prediction error based on the gradient of the error concerning the prediction. After parameter tuning, we found that a max tree depth of 3 produced the best accuracy at 82.0\%. The precision of the model was 15.3\%, negative predictive value was high at 97.4\%, sensitivity was 57.0\%, and specificity was 83.4\%. F1 score for this algorithm was 24.1\%.
\begin{table*}[t]
    \centering
    \begin{tabular}{|c|c|c|c|c|c|c|c|}
    \hline
    \multicolumn{8}{|c|}{\textbf{Practice Fusion Dataset}}                                                                                                                                                                                                                            \\ \hline
    \textbf{Classifier- Performance} & \textbf{K-NN}                          & \textbf{SVM} & \textbf{Decision Tree} & \textbf{Random Forest}                 & \textbf{Gradient Boosting} & \textbf{Neural Network}                & \textbf{Naive Bayes}                   \\ \hline
    \textbf{Accuracy}                & 0.8151                                 & 0.8114       & 0.7923                 & 0.8238                                 & 0.8204                     & {\color[HTML]{009901} \textbf{0.8254}} & 0.6643                                 \\ \hline
    \textbf{Precision}               & 0.0610                                 & 0.0054       & 0.2352                 & 0.1239                                 & 0.1526                     & 0.1831                                 & {\color[HTML]{009901} \textbf{0.6643}} \\ \hline
    \textbf{Negative Prediction}     & {\color[HTML]{009901} \textbf{0.9881}} & 0.9963       & 0.9201                 & 0.9843                                 & 0.9736                     & 0.9728                                 & 0.6433                                 \\ \hline
    \textbf{Sensitivity}             & 0.5396                                 & 0.2500       & 0.4031                 & {\color[HTML]{009901} \textbf{0.6449}} & 0.5705                     & 0.6071                                 & 0.3122                                 \\ \hline
    \textbf{Specificity}             & 0.8210                                 & 0.8137       & 0.8399                 & 0.8304                                 & 0.8336                     & 0.8385                                 & {\color[HTML]{009901} \textbf{0.8961}} \\ \hline
    \textbf{F1-Score}                & 0.1097                                 & 0.0105       & 0.2971                 & 0.2078                                 & 0.2408                     & 0.2814                                 & {\color[HTML]{009901} \textbf{0.4248}} \\ \hline
    \end{tabular}
    \caption{Overall Performance of Individual Classifiers for Practice Fusion Dataset}
    \label{tab:class_perform_practice}
\end{table*}

\begin{table*}[t]
    \centering
        \begin{tabular}{|c|c|c|c|c|c|c|c|}
    \hline
    \multicolumn{8}{|c|}{\textbf{Pima Indians Dataset}}                                                                                                                                                                                                            \\ \hline
    \textbf{Classifier- Performance} & \textbf{K-NN}                          & \textbf{SVM} & \textbf{Decision Tree}                 & \textbf{Random Forest}        & \textbf{Gradient Boosting} & \textbf{Neural Network}       & \textbf{Naive Bayes}          \\ \hline
    \textbf{Accuracy}                & 0.6941                                 & 0.3427       & {\color[HTML]{009901} \textbf{0.7260}} & {\color[HTML]{000000} 0.7123} & 0.6849                     & {\color[HTML]{333333} 0.3425} & 0.6803                        \\ \hline
    \textbf{Precision}               & 0.2800                                 & 1.000        & {\color[HTML]{009901} \textbf{0.6400}} & 0.5067                        & 0.3733                     & 1.000                         & {\color[HTML]{333333} 0.3467} \\ \hline
    \textbf{Negative Prediction}     & {\color[HTML]{009901} \textbf{0.9097}} & 0.000        & 0.7708                                 & 0.8194                        & 0.8472                     & 0.000                         & 0.8542                        \\ \hline
    \textbf{Sensitivity}             & {\color[HTML]{009901} \textbf{0.6176}} & 0.3425       & 0.5926                                 & {\color[HTML]{000000} 0.5938} & 0.5600                     & 0.3425                        & 0.5532                        \\ \hline
    \textbf{Specificity}             & 0.7081                                 & 0.635        & {\color[HTML]{009901} \textbf{0.8043}} & 0.7613                        & 0.7219                     & 0.7152                        & {\color[HTML]{000000} 0.7151} \\ \hline
    \textbf{F1-Score}                & 0.3853                                 & 0.0105       & {\color[HTML]{009901} \textbf{0.6154}} & 0.5468                        & 0.4480                     & 0.5102                        & {\color[HTML]{000000} 0.4262} \\ \hline
    \end{tabular}
    \caption{Overall Performance of Individual Classifiers for Pima Indians Dataset}
    \label{tab:class_perform_pima}
\end{table*}

\subsection{Neural Network}
For our neural network algorithm, we used Multi-Layer Perceptron Neural Networks(MLP). Neural networks\cite{huang1988neural} are inspired by the way our human biological neural networks in the brain works. It consists of neurons, weights, and activation or transfer functions. A perceptron is a single neuron model. Think of a neuron as a unit that receives input. Each input the neuron receives is given a weight, and as it passes through the neuron, the neuron or node applies the activation function to the sum of the weighted inputs. Each activation function performs a fixed mathematical operation on the sum of the inputs. Multiple neurons are then connected in both the input layer, hidden layer(s), and the output layer. A Multi-Layer Perceptron Neural Network has one or more hidden layers that can perform non-linear functions. MLP trains on data to model the correlation between inputs and outputs by a constant back and forth pass with the output layer being compared against the actual true classification going forward. And then the backward propagation with partial derivatives of the error function with regards to various weights to move the MLP closer and closer to error minimum. So inputs are moving forward from one neuron to another, and multiple functions are being applied to the sum of the inputs as it goes forward. But it is also constantly evaluating and going backward to adjusted weights and biases until it reaches optimum minimum error and a final classification decision. Data was first scaled as neural networks like this with deep learning expect all input variables to vary in the same vary or similar way. Scaling the data will produce a variance of 1 for all variables and mean of 0. After that, we ran the algorithm and found that accuracy was about 82.5\%. The precision of the model was 18.3\%, negative predictive value was high at 97.3\%, sensitivity was 60.7\%, and specificity was 83.8\%. F1 score for this algorithm was 28.1\%.

\subsection{Naive Bayes}
\begin{figure}[h!]
    \centering
    \includegraphics[scale=0.35]{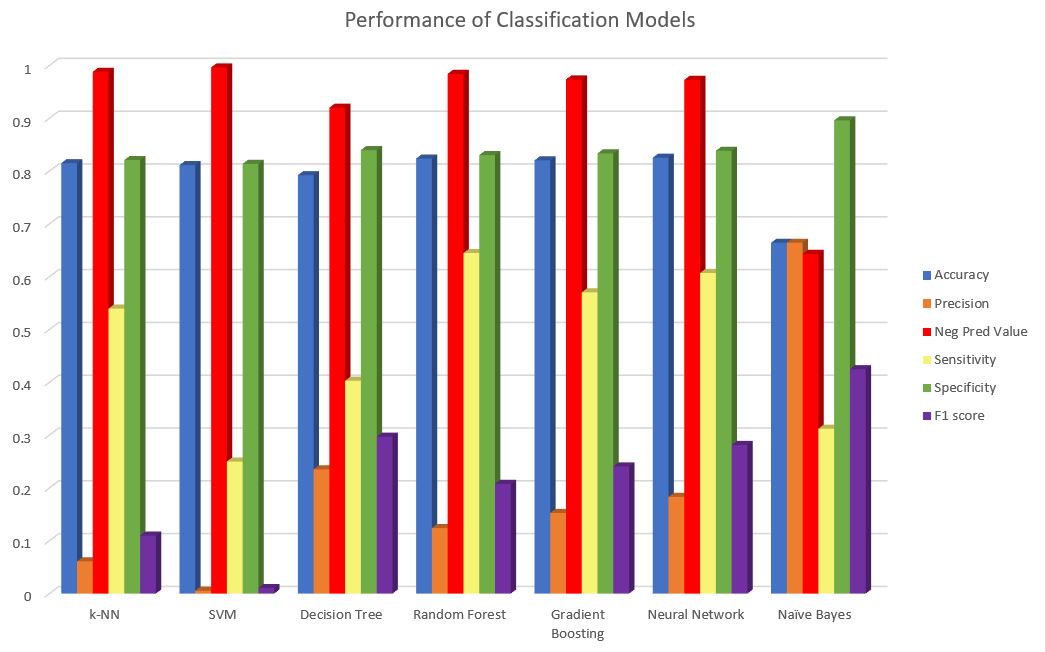}
    \caption{Classification Performance of Machine Learning Algorithms for Practice Fusion Dataset}
    \label{fig:tabtochart_practice}
\end{figure}

\begin{figure}[h!]
    \centering
    \includegraphics[scale=0.5]{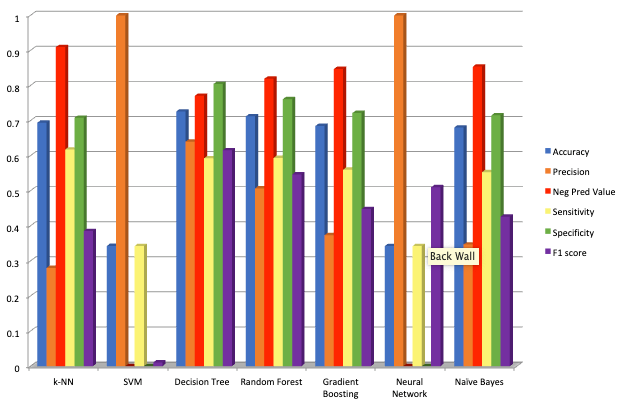}
    \caption{Classification Performance of Machine Learning Algorithms Pima Indians Dataset}
    \label{fig:tabtochart_pima}
\end{figure}

Naive Bayes is a probabilistic classification algorithm that that is based on Bayes’ Theorem. Bayesian classifiers assign the most likely class to a given sample described by its feature vector. The naive Bayes classifier\cite{rish2001empirical} greatly simplifies learning by assuming that features are independent of given class, which can be formulated as \[P(X|C) = \prod_i=1^n P(X_i|C), where X = (X_1, X_2, ......, X_n) \]. It works on conditional probability, the probability that something will occur given that something else has already occurred. Naive Bayes assumes that every variable being classified is independent of any other variable i.e., predictors are independent. So, this algorithm considers each variable to contribute independently to the probability of the classification being either diabetic or non-diabetic. The algorithm will calculate the posterior probability for each class using the Naive Bayesian equation, and the class with high posterior probability will be the outcome. This is fairly simplistic compared to some of the other models that we have tried, but sometimes a simpler model may work better. In this case, however, it did not. Accuracy for this algorithm was the lowest of all models at 66.4\%. The precision of the model was higher at 66.4\%, but the negative predictive value was much lower at 66.4\%. Sensitivity was 31.2\%, and specificity was 89.6\%. F1 score for this algorithm was 42.4\%. Performance of individual classifiers for Pima Indian and Practice Fusion datasets are shown in \ref{fig:tabtochart_pima}, \ref{tab:class_perform_pima} and \ref{fig:tabtochart_practice}, \ref{tab:class_perform_practice} respectively. As from the Table \ref{tab:class_perform_practice}, we can observe that most classification algorithms performed well with about 80-82\% accuracy other than the Naive Bayes. However, all algorithms other than Naive Bayes also suffered from a very low precision meaning that of the patients that are actually classified as diabetic, there is a very low chance of them being diabetic. That means that there are a lot of false positives being reported. Conversely, the negative predictive value for the classifiers is all very high and consistently higher than 90\% meaning that when the algorithms classify a patient as non-diabetic, they are very sure that the patient is actually non-diabetic.

 \subsection{Ensemble Model}
Always, it is better for us to have more false positives in this case than false negatives because as far as diabetes detection goes, we rather a patient get tested than ignore testing thinking they do not have diabetes. So we will take it one step further by implementing a Weighted Average Ensemble model. Of the classification algorithms above, we will create an ensemble model that will take the result of each algorithm in the model, multiply it by the weight assigned to the algorithm and then sum the result to determine the final classification outcome as shown in Fig \ref{fig:block}. 
\begin{table*}[t]
    \centering
    \begin{tabular}{|c|c|c|}
    \hline
    \textbf{Classifier- Performance} & \textbf{Practice Fusion}              & \textbf{Pima Indians}                 \\ \hline
    \textbf{Accuracy}                & {\color[HTML]{333333} 0.860}          & {\color[HTML]{009901} \textbf{0.891}} \\ \hline
    \textbf{Precision}               & 0.242                                 & {\color[HTML]{009901} \textbf{0.769}} \\ \hline
    \textbf{Negative Prediction}     & {\color[HTML]{009901} \textbf{0.992}} & 0.958                                 \\ \hline
    \textbf{Sensitivity}             & 0.884                                 & {\color[HTML]{009901} \textbf{0.909}} \\ \hline
    \textbf{Specificity}             & {\color[HTML]{000000} 0.847}          & {\color[HTML]{009901} \textbf{0.884}} \\ \hline
    \textbf{F1-Score}                & 0.380                                 & {\color[HTML]{009901} \textbf{0.833}} \\ \hline
    \end{tabular}
    \caption{Overall Performance of Ensemble Model on Practice Fusion and Pima Indian Datasets}
    \label{tab:ensemble_perform}
\end{table*}
To implement this, we first select the models that we like to combine and then we run an optimization model to see what weights to assign to each model to produce the highest final accuracy. First, we take all models except for Naive Bayes, and after running them through optimization for log loss, we find that optimal ensemble model involve a combined model of SVM, Random Forest, and Gradient Boosting for Practice Fusion dataset and a combined model of KNN, Random Forest, and Gradient Boosting for Pima Indian dataset. The optimized weights of both these datasets are 0.073, 0.902, 0.023 and 3.262, 5.431, 5.453 respectively.  After running the combined ensemble consisting of these three algorithms with weights, the accuracy did increase to 85\%. Precision was  24.2\%, Negative predictive value 99.3\%, Sensitivity 88.5\%, and Specificity 84.7\%. F1 score was 38.0\%. Overall performance of the Ensemble on Practice Fusion and Pima Indians are shown in \ref{tab:ensemble_perform}.

\section{CONCLUSION and FUTURE WORK}
The supervised ensemble model in this work will help determine whether a patient may have diabetes or at high risk of developing diabetes. All algorithms other than Naive Bayes also suffered from a very low precision meaning that of the patients that are actually classified as diabetic, there is a very low chance of them being diabetic. That means that there are a lot of false positives being reported. Conversely, the negative predictive value for the classifiers is all very high and consistently higher than 90\% meaning that when the algorithms classify a patient as non-diabetic, they are very, very sure that the patient is non-diabetic. Weighted Average Ensemble model performed better in recovering wrong predictions. Unlike the previous works that focused either particular classifier-set or a Pima Indians dataset that is heavily biased towards limited female population, we use a new approach and dataset, yet use the Pima Indians dataset for the baseline comparison. While the accuracy of the previous works on Pima Indians dataset was less than 80\%, the accuracy of our Ensemble model reached 89\% for the same dataset. The accuracy of the Ensemble model on Practice Fusion is 85\%, by far our ensemble approach is new in this space. We firmly believe that the weighted average ensemble model not only performed well in overall metrics but also helped in recover wrong predictions and aid in accurate prediction of Type2 diabetes. Our accurate novel model can be used as an alert for the patients to seek medical evaluation in time. Clinicians could use the tool to determine whether they should test a patient earlier for diabetes with the industry standard HBA1c\cite{rohlfing2000use}test, which measures glycated hemoglobin and long-term average plasma glucose concentration. The model could ultimately be made available for direct patient use with a more limited set of biometrics and be used as an alert to patients to seek medical evaluation in time. Similarly, this decision support system may also be implemented by larger health entities like health care systems or public health systems with more biometrics to quickly look at a large population and see how many people may have diabetes and decide on larger long-term health implementations from there.

\bibliographystyle{IEEEtran}
\bibliography{bibliography}

\end{document}